\documentclass[fleqn,10pt,twocolumn]{wlscirep}
\usepackage[utf8]{inputenc}
\usepackage[T1]{fontenc}
\title{Mobile CT Services for Rural, Regional, and Remote Areas: Current Practice and Future Integration with Telehealth and Regulatory-Authorised AI}

\makeatletter
\newcommand\setcurrentname[1]{\def\@currentlabelname{#1}}
\makeatother
\usepackage{subcaption}
\usepackage{makecell}
\usepackage{multirow}
\usepackage{multicol}
\usepackage{geometry}
\newcommand{\PreserveBackslash}[1]{\let\temp=\\#1\let\\=\temp}
\newcolumntype{C}[1]{>{\PreserveBackslash\centering}m{#1}}

\usepackage{hyperref}
\usepackage{xr-hyper}
\usepackage{lineno}
\usepackage{pdflscape}
\usepackage{longtable}
\usepackage{fontawesome5}
\definecolor{darkgreen}{RGB}{0,200,0} 
\usepackage[most]{tcolorbox}

\newtcolorbox{casestudybox}{
    colback=gray!5,
    colframe=black!70,
    boxrule=0.7pt,
    arc=1.5mm,
    left=2mm,
    right=2mm,
    top=1.5mm,
    bottom=1.5mm,
    before skip=8pt,
    after skip=8pt,
    fonttitle=\bfseries,
    title={Case Study: Mobile CT, Telehealth and AI in Western China}
}

\author[1]{Zhicheng Lu}
\author[2]{Md Zahid Islam}
\author[3]{M Mamun Huda}
\author[4]{Kristie Sweeney}
\author[5]{Shayne Chau}
\author[4]{Oliver Mulcock}
\author[4]{Corey Hemopo}
\author[3]{Catherine Keniry}
\author[1,2,6*]{Mohammad Ali Moni}

\affil[1]{Rural Health Research Institute, Charles Sturt University, Orange, NSW 2800, Australia}
\affil[2]{Artificial Intelligence and Cyber Futures Institute, Charles Sturt University, Bathurst, NSW 2795, Australia}
\affil[3]{School of Rural Medicine, Charles Sturt University, Orange, NSW 2800, Australia}
\affil[4]{Western NSW Local Health District, Bathurst, NSW 2795, Australia}
\affil[5]{School of Dentistry and Medical Sciences, Charles Sturt University, NSW 2650, Australia}
\affil[6]{School of Health and Rehabilitation Sciences, The University of Queensland, St Lucia, Brisbane, QLD 4072, Australia}
\affil[*]{Corresponding author: m.moni@uq.edu.au}

\begin{abstract}
Computed tomography (CT) plays an essential role in clinical workflow to improve patient outcomes. However, access to CT imaging and specialist interpretation remains limited, particularly in rural, regional, remote (RRR), and other resource-limited settings. Recent advances in mobile CT, telehealth, and artificial intelligence (AI) provide opportunities to extend advanced imaging services to populations in RRR settings. This review examines: 1) mobile CT systems deployed in trucks, trailers, ambulances, and other mobile platforms; 2) telehealth technologies supporting CT-based healthcare; and 3) AI for CT that has received regulatory authorisation or is currently deployed in clinical practice. Applications are evaluated across four clinical functions: screening and diagnosis, patient monitoring, risk prediction, and intervention or therapeutic decision support. The review covers neurological, thoracic, cardiovascular, abdominal, oncological, musculoskeletal, and interventional imaging, with particular attention to stroke, cancer, and other image-guided treatment. Other factors such as regulatory status, deployment status, and estimated technology readiness (TRL) level are compared. Current evidence indicates that mobile CT, telehealth, and AI for conventional CT are individually relatively mature, but fully integration of these technologies remains less widely deployed and validated in the clinical settings. Key barriers include regulatory variation, domain shift, connectivity requirements, cost, workflow integration, cybersecurity, and limited evidence of patient-level benefit. Future research should prioritise prospective, multicentre evaluation of integrated CT systems in real-world and underserved clinical settings.
\end{abstract}
\begin{document}

\flushbottom
\maketitle
%
%
\thispagestyle{empty}

\section{Introduction}

\begin{figure*}
    \centering
    \includegraphics[width=\linewidth]{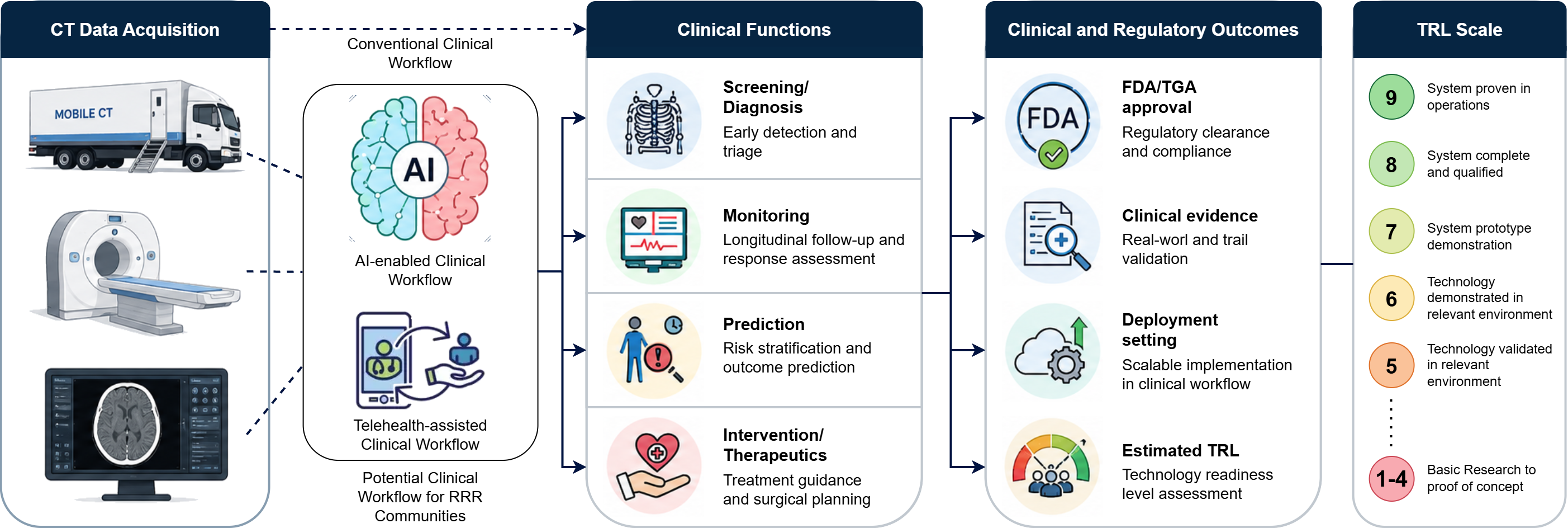}
    \caption{Overview of the review framework for mobile and conventional CT integrated with AI and telehealth, covering four clinical functions: screening and diagnosis, patient monitoring, prediction, and intervention or therapeutic decision support. We also evaluate them according to regulatory status, clinical deployment, and estimated TRL.}
    \label{fig:classification_framework}
\end{figure*}

Computed tomography (CT) is an essential component of healthcare. Its fast acquisition time, wide availability compared with magnetic resonance imaging (MRI), high spatial resolution, and ability to visualise important anatomical structures (e.g., bone, soft tissue, vasculature, and acute pathological changes) make it particularly important in time-critical conditions. CT is one of the gold-standard imaging modalities for suspected acute stroke, traumatic injury, pulmonary embolism, cancers, numerous oncological conditions, and many other surgical applications \cite{macintosh2023radiological}. More broadly, diagnostic technologies influence clinical decision-making throughout the clinical pathway, from initial disease detection and diagnosis to treatment planning, therapeutic monitoring, and follow-up \cite{WHO2026Diagnostics}. However, access to advanced CT imaging may still become limited because conventional CT systems are expensive and require substantial infrastructure, trained radiographers, regular maintenance, and specialist interpretation. This situation is more serious in rural, regional, remote (RRR), and other resource-constrained settings, where even less imaging equipment and qualified healthcare professionals further restrict access to diagnostic services \cite{Frija2021ImagingAccess,WHO2025ImagingResolution}. For example, in Australia around 7 million Australians (27\% of the population) live in rural or remote areas, and 58\% of those have limited access to medical imaging and specialists, compared with only 6\% in urban areas \cite{Welfare2018Australian,Welfare2025Australian}. This disparity makes equitable access to medical imaging services more critical.

To address these challenges, mobile CT systems have been introduced and developed to bring imaging services to patients in RRR areas rather than requiring those patients to travel hours for imaging services. Rather than portable bedside CT scanners designed primarily for intensive care units or operating rooms, this review focuses on mobile CT systems deployed in trucks, trailers, or other mobile platforms to extend imaging services to RRR and other underserved communities. Such systems may improve access to diagnostic imaging, reduce delays in diagnosis, and assist earlier clinical decision-making where permanent imaging infrastructure is not available locally.
Mobile CT has also played a critical role in neurological emergencies through mobile stroke units (MSUs), which integrate CT imaging and specialised personnel within a mobile ambulance-based service \cite{Walter2022ESO}. In a prospective multicentre controlled trial, MSU care was associated with earlier thrombolytic treatment and improved functional outcomes compared with conventional medical services \cite{Grotta2021MSU}. Nevertheless, the wider deployment of mobile CT remains constrained by infrastructure requirements, operational and maintenance costs, and regulatory requirements \cite{Walter2022ESO}.

Telehealth provides another key component by enabling images and associated clinical information to be securely transmitted to remotely located (e.g., urban areas) specialists. Through secure image transfer, remote interpretation, and video consultation, telehealth can allow patients in RRR areas to receive specialist support without immediate and long-time transfer. In acute stroke care, telehealth and telestroke networks can support CT interpretation and diagnostics, treatment decisions, and potential transfer planning between primary and comprehensive stroke centres \cite{Walter2022ESO}. Telehealth can also support remote radiology reporting and specialist consultation for suspected conditions and regular health check. However, its effectiveness and real-world deployment depend heavily on reliable connectivity and cybersecurity issues.

\begin{table*}[t]
    \centering
    \caption{Information extracted from each technology}
    \label{tab:information_extracted}
    \begin{tabular}{c|c}
        \hline
        \textbf{Category} & \textbf{Information Collected} \\
        \hline
        Basic information & Product name, manufacturer, country \\
        \hline
        Technology & Mobile CT / AI / Telehealth \\
        \hline
        Clinical application & Stroke, lung, liver, trauma, etc. \\
        \hline
        Function & Screening, monitoring, prediction, intervention \\
        \hline
        Regulatory status & FDA, TGA, CE, etc. \\
        \hline
        Clinical deployment & Commercial, pilot, research \\
        \hline
        Evidence & Prospective, retrospective, multicentre \\
        \hline
        Interoperability & PACS, cloud, HIS, telehealth \\
        \hline
        Estimated TRL & 6-9 \\
        \hline
        Advantages & Key strengths \\
        \hline
        Limitations & Current challenges \\
        \hline
    \end{tabular}
\end{table*}

\begin{figure*}[t]
    \centering
    \includegraphics[width=.7\linewidth]{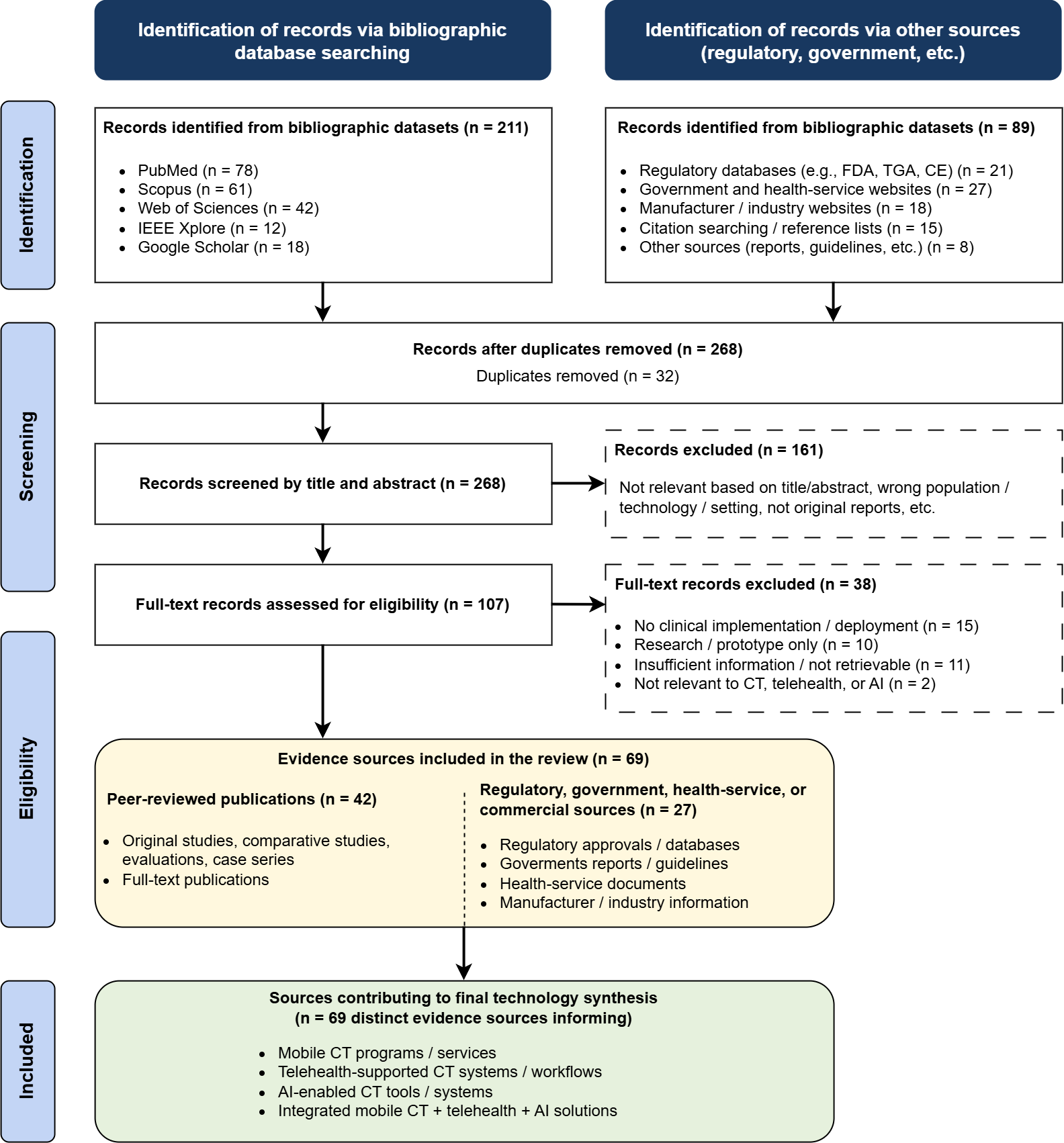}
    \caption{PRISMA 2020 flow diagram of evidence identification and selection. A total of 211 records were identified through bibliographic database searching and 89 additional records were identified through regulatory databases, government and health-service websites, manufacturer sources, citation searching, and other sources. After removal of 32 duplicates, 268 records were screened by title and abstract, 107 were assessed in full text, and 69 evidence sources (42 peer-reviewed publications and 27 from other sources) were included in the final technology synthesis.}
    \label{fig:PRISMA}
\end{figure*}

Artificial intelligence (AI) has emerged as the third important technology for improving CT-based healthcare. Recent advances in machine learning and deep learning have enabled AI systems to assist with image reconstruction, lesion detection, organ and lesion segmentation, quantitative measurement, and thus clinical decision support \cite{Singh2025FDAAI}. Numerous AI-enabled medical devices have received marketing authorisation from the United States Food and Drug Administration (FDA), who maintains a public list of authorised AI-enabled devices \cite{FDA2026AIList}. In Australia, AI software that meets the definition of a medical device is regulated according to its intended purpose and associated risk and generally must be included in the Australian Register of Therapeutic Goods (ARTG) before it can be legally supplied \cite{TGA2026AIRegulation,TGA2025AIARTG}. These systems are generally intended to support, rather than replace, clinicians by providing alerts, quantitative measurements, or decision-support information. Although many CT-based AI systems have received regulatory authorisation, regulatory approval does not necessarily guarantee generalisability across different patient populations, institutions, scanner models, or clinical environments \cite{Windecker2025Generalizability}. International guidelines therefore emphasise rigorous external validation, real-world evaluation, and continuous monitoring to ensure the safe and trustworthy deployment of AI in healthcare \cite{Lekadir2025FutureAI}. More recently, telehealth platforms have been combined with AI-assisted CT analysis to provide automated alerts and facilitate rapid communication between patients in RRR settings and healthcare provider teams. In these workflows, AI may identify or prioritise a suspected time-critical finding, while communication software distributes images and notifications to relevant specialists. This integration has the potential to reduce delays between image acquisition, specialist review, patient transfer, and treatment. Evaluations of integrated systems should therefore extend beyond diagnostic accuracy to real-world clinical performance \cite{Vasey2022DECIDEAI,Lekadir2025FutureAI}.

Although mobile CT, telehealth and AI have each developed rapidly, they are frequently evaluated as separate areas. Existing mobile CT studies commonly compare feasibility, mobility, image quality, and other factors with conventional CT. Telehealth reviews principally examine remote communication and service delivery, while AI reviews often focus on algorithmic performance (e.g., detection rate or segmentation Dice). Those pieces do not fully represent emerging clinical systems in which mobile CT, image transfer, AI analysis, specialist interpretation, and treatment decision-making operate as an integrated and complete pathway. It may also lead to conflating distinct technologies, such as a portable CT scanner, a smartphone application displaying AI-generated alerts from a fixed CT scanner, and a complete mobile clinical service that includes CT, telehealth, and decision-support capabilities.

Therefore, this review provides a comprehensive overview of three complementary technologies for improving access to CT-based healthcare: (1) mobile CT systems deployed to support healthcare delivery in RRR and underserved communities; (2) telehealth technologies supporting CT-based diagnosis and patient management; and (3) AI software for CT that has received regulatory authorisation or has been implemented in clinical practice. These technologies are examined across four major clinical functions: screening and diagnosis, patient monitoring, risk prediction, and intervention or therapeutic decision support. We further compare their regulatory status, deployment status, estimated technology readiness levels (TRLs), advantages, limitations, and opportunities for integration. By examining mobile CT, telehealth, and AI within a unified framework (see Figure \ref{fig:classification_framework}), this review aims to provide a practical reference for researchers, clinicians, healthcare providers, industry partners, and policy-makers seeking to improve equitable access to advanced CT services.

\section{Review Methodology}

This study was conducted as a structured scoping technology review integrating peer-reviewed evidence with regulatory records, government and health-service documentation, manufacturer information, and publicly available evidence of clinical deployment. The review was designed to identify technologies that had progressed beyond proof-of-concept development and to compare their current clinical maturity, deployment characteristics, and potential for integration within distributed and mobile CT services.

\subsection{Literature Search Strategy}

A comprehensive search was conducted to identify commercially available, regulatory-authorised, and clinically deployed technologies related to mobile CT, telehealth, and AI. Searches were performed in PubMed, Scopus, Web of Science, IEEE Xplore, and Google Scholar to the date of 31$^\text{st}$, August, 2026.

Additional searches were undertaken using regulatory and government sources, including the United States FDA, the Australian Therapeutic Goods Administration (TGA), ARTG, the United Kingdom National Institute for Health and Care Excellence (NICE), the China National Medical Products Administration (NMPA), and relevant national or regional health-service websites. Manufacturer documentation was used where necessary to confirm commercial availability, intended use, interoperability, or deployment status \cite{FDA2026AIList,TGA2026AIRegulation}. Searches were also performed on 31$^\text{st}$, August, 2026.

Search strategies combined terms related to CT, mobile imaging, AI, telehealth, clinical deployment, and regulatory authorisation. Core search concepts included ``computed tomography'', ``CT'', ``mobile CT'', ``mobile computed tomography'', ``portable CT'', ``mobile stroke unit'', ``artificial intelligence'', ``deep learning'', ``machine learning'', ``computer-aided diagnosis'', ``telehealth'', ``telemedicine'', ``teleradiology'', ``telestroke'', ``FDA'', ``510(k)'', ``De Novo'', ``TGA'', ``ARTG'', ``NICE'', ``NMPA'', ``clinical deployment'', and ``clinical implementation''. Search terms were combined using Boolean operators and adapted to the syntax of each database. Full database-specific search strategies are provided in Supplementary Table S1.

Reference lists of relevant reviews and key primary studies were manually screened to identify additional eligible publications and technologies. Forward citation searching was also performed for selected key studies and established mobile CT programs where appropriate.

\subsection{Study Selection and Screening}

Duplicate records were removed before screening. Titles and abstracts were screened for relevance, followed by full-text assessment of potentially eligible publications.

A total of 211 records were identified through bibliographic database searching, with an additional 89 records identified through regulatory databases, government and health-service websites, manufacturer sources, citation searching, and other grey-literature sources. After removal of 32 duplicates, 268 records underwent title and abstract screening, of which 107 were assessed in full text. Ultimately, 42 peer-reviewed publications and 27 regulatory, government, health-service, or commercial sources contributed to the final technology synthesis.

Screening was performed by 2 reviewers. Where more than one reviewer was involved, disagreements regarding eligibility or classification were resolved through discussion. The study-selection process is summarised in Figure \ref{fig:PRISMA}.

\subsection{Eligibility Criteria}

The primary objective of this review was to evaluate technologies that had progressed beyond proof-of-concept research towards clinical implementation. Technologies or programs were included if they satisfied at least one of the following criteria:

\begin{itemize}
    \item commercially available or operational mobile CT systems currently used in clinical practice, screening, emergency care, or health-service delivery;
    \item telehealth, teleradiology, telestroke, or related clinical systems routinely used to support CT-based diagnosis, specialist interpretation, patient management, or treatment decision-making;
    \item CT-related AI software that had received regulatory authorisation, including FDA clearance or approval, inclusion in the ARTG, CE-marked deployment where adequately documented, NMPA approval, or other recognised jurisdiction-specific authorisation;
    \item CT-related AI software without clearly documented regulatory authorisation where there was sufficient evidence of current clinical deployment;
    \item integrated systems combining two or more of mobile CT, telehealth, and AI with documented clinical implementation or prospective clinical evaluation.
\end{itemize}

Research prototypes without clinical validation, purely retrospective algorithm-development studies, simulation-only investigations, conference abstracts, and technologies lacking evidence of clinical implementation were excluded. Exceptions were made only for emerging integrated technologies with substantial translational relevance to mobile or distributed CT pathways, provided that their developmental status was explicitly identified. Portable or bedside CT scanners were primarily excluded unless they formed part of a mobile or transportable clinical service relevant to the aims of this review.

\subsection{Technology Classification Framework}

Eligible systems were categorised into three major groups according to their primary role within the CT clinical pathway:

\begin{enumerate}
    \item Mobile CT systems;
    \item Telehealth systems supporting CT-based healthcare;
    \item AI for CT.
\end{enumerate}

The overall organisation of this review and the relationship between these technologies and their clinical applications are illustrated in Figure~\ref{fig:classification_framework}.

Mobile CT deployments were additionally classified according to their primary service model, including general diagnostic mobile CT, mobile screening services, mobile stroke units, relocatable or continuity-of-service CT, and deployable or austere-environment CT.

Within each technology category, systems were further classified according to four principal clinical functions:

\begin{itemize}
    \item screening and diagnosis;
    \item patient monitoring;
    \item risk prediction;
    \item intervention and therapeutic decision support.
\end{itemize}

For each technology or service, information was extracted where available on the manufacturer or developer, country or deployment region, technology type, intended clinical application, target disease, regulatory status, clinical deployment status, advantages, limitations, and estimated TRL. The information collected from each included technology is summarised in Table~\ref{tab:information_extracted}.

For mobile CT services, operational deployment was distinguished from temporary demonstrations or prototype use. For AI systems, regulatory authorisation was considered separately from evidence of routine clinical deployment because regulatory clearance alone does not establish real-world effectiveness across different institutions, patient populations, scanner configurations, or deployment environments.

\subsection{Technology Readiness Assessment}

Because formal TRL values were rarely explicitly reported, TRL was estimated by the authors using published clinical evidence, regulatory status, commercial availability, and degree of documented operational deployment \cite{Mankins1995TRL,Heder2017TRL}. The following definitions were used:

\begin{itemize}
    \item \textbf{TRL 6.} Technology demonstrated in a relevant clinical environment;
    \item \textbf{TRL 7.} Prototype or integrated system demonstrated in an operational clinical environment;
    \item \textbf{TRL 8.} Complete system clinically validated, qualified, regulatory-authorised, or otherwise demonstrated as ready for operational deployment;
    \item \textbf{TRL 9.} Complete system demonstrated through routine or established operational clinical use.
\end{itemize}

Where evidence supported an intermediate classification, a range such as TRL 8--9 was assigned. TRL estimates were intended as comparative indicators of technology and deployment maturity rather than formal manufacturer-assigned readiness levels. Importantly, component-level maturity was not assumed to transfer directly to an integrated system. For example, an AI system operating at TRL 9 in conventional hospital CT was not automatically considered TRL 9 when transferred to a mobile CT environment without separate validation.

TRL assessments were performed by 2 authors. Where two authors independently assessed TRL, disagreements were resolved by discussion or averaging.






\subsection{Synthesis Approach}

The review primarily focuses on practical clinical implementation rather than algorithmic performance alone. Consequently, greater emphasis was placed on regulatory-authorised technologies, routinely deployed services, prospective clinical evaluation, and documented real-world implementation than on experimental AI models evaluated solely using retrospective datasets. Findings were therefore synthesised narratively and through comparative tables according to technology type, clinical function, regulatory status, deployment setting, and estimated TRL. Particular attention was given to the maturity gap between individually established technologies and their integration into end-to-end mobile CT, telehealth, and AI clinical pathways.

\begin{table*}[pt]
\centering
\small
\setlength{\tabcolsep}{3pt}
\renewcommand{\arraystretch}{1.1}

\caption{Operational and clinically deployed mobile CT programs and services. \textcolor{darkgreen}{\faTruck}: mobile CT is operational. \textcolor{darkgreen}{\faPhone}: telehealth is operational. \textcolor{orange}{\faPhone}: telehealth is partially operational. \textcolor{darkgreen}{\faMicrochip}: AI is operational. \textcolor{orange}{\faMicrochip}: AI is partially operational}
\label{tab:mobile_CT}

\makebox[\textwidth][c]{%
\begin{tabular}{
    C{1.6cm}|
    C{3cm}|
    C{8cm}|
    C{2cm}|
    C{1.5cm}|
    C{1.5cm}
}
\hline
Country & Program & Description & Source & Technology & Estimated TRL \\
\hline

Argentina & Buenos Aires MSU & Operational Mobile Stroke Unit providing prehospital CT imaging. & Kim \textit{et al.} \cite{kim2024global} & \textcolor{darkgreen}{\faTruck} \textcolor{darkgreen}{\faPhone} & 9 \\
\hline

\multirow[c]{4}{1.6cm}[-8em]{\centering Australia} & Western NSW Remote Mobile CT Service & Full-size CT installed in a purpose-built truck serving Bourke, Walgett, Cobar and surrounding rural communities. Images are uploaded into the NSW imaging network for reporting. Telehealth consultation and AI are not explicitly reported. & \href{https://www.ehealth.nsw.gov.au/news/2023/mobile-ct-service-in-rural-nsw}{eHealth NSW}\cite{eHealth2026Mobile} & \textcolor{darkgreen}{\faTruck} & 9 \\
\cline{2-6}
 & Heart of Australia HEART 5 & Battery-powered mobile CT truck providing respiratory and general CT services to rural and remote Queensland. Commercial clinical service. No published AI or embedded telehealth workflow identified. & \href{https://heartofaustralia.com.au/2024/10/heart-of-australia-announces-nationwide-expansion}{Heart of Australia} \cite{Heart2025Heart} & \textcolor{darkgreen}{\faTruck} & 9 \\
\cline{2-6}
 & National Lung Cancer Screening Program (HEART 7–11) & Australian Government programme deploying five mobile CT trucks across WA, NT, QLD, NSW/VIC/TAS and SA to provide LDCT screening in rural and remote communities. Results are returned to the referring GP. AI has not been reported. & \href{https://www.health.gov.au/our-work/nlcsp/how-it-works?language=en}{Department of Health, Disability and Ageing} \cite{Health2026How} & \textcolor{darkgreen}{\faTruck} \textcolor{orange}{\faPhone} & 9 \\
\cline{2-6}
 & Australian Defence Force Deployable CT Capability & A deployable full diagnostic CT scanner was used during a field exercise in 2026. The system supports trauma and operational medicine in transportable and austere environments. It is an operational military capability rather than a routine civilian rural service. & \href{https://www.defence.gov.au/news-events/news/2026-06-04/ct-scanner-game-changer-medics-field}{Department of Defence} \cite{Defence2026CT} & \textcolor{darkgreen}{\faTruck} & 8-9 \\
\cline{2-6}
 & Melbourne Mobile Stroke Unit & First MSU in the Southern Hemisphere. Built-in CT scanner, telemedicine equipment, mobile laboratory, neurologist, stroke nurse, CT radiographer and paramedics. & \href{https://strokefoundation.org.au/what-we-do/research/stroke-ambulance}{Stroke Foundation} \cite{Stroke2025Stroke} & \textcolor{darkgreen}{\faTruck} \textcolor{orange}{\faPhone} & 9 \\
\hline

\multirow[c]{2}{1.6cm}[-1em]{\centering Brazil} & Barretos Cancer Hospital Mobile Lung Screening & Mobile LDCT programme providing screening and follow-up for underserved populations. No explicit telemedicine or AI workflow reported. & Chiarantano \textit{et al.}, 2022 \cite{chiarantano2022implementation} & \textcolor{darkgreen}{\faTruck} & 9 \\
\cline{2-6}
 & Brasília MSU & Operational Mobile Stroke Unit identified in the PRESTO registry. & Kim \textit{et al.} & \textcolor{darkgreen}{\faTruck} \textcolor{darkgreen}{\faPhone} & 9 \\
\hline

\multirow[c]{2}{1.6cm}[-2.5em]{\centering Canada} & Canadian Mobile CT Units (British Columbia \& Quebec) & CADTH/CMII reports three publicly funded mobile CT trailers, serving at least 20 municipalities and First Nations. Two are in British Columbia and one in Quebec. & CADTH \cite{CADTH2021Canadian} & \textcolor{darkgreen}{\faTruck} & 9 \\
\cline{2-6}
 & PHSA Mobile CT Trailer & Provincial mobile CT trailer supporting hospitals during planned or unplanned CT downtime. Primarily a continuity-of-service model rather than rural outreach. & \href{https://www.phsa.ca/our-services/programs-services/provincial-medical-imaging-office}{PHSA} \cite{PHSA2026Provincial} & \textcolor{darkgreen}{\faTruck} & 9 \\
\cline{2-6}
 & Edmonton Stroke Ambulance & CT-equipped stroke ambulance providing prehospital diagnosis and thrombolysis. & Kim \textit{et al.} \cite{kim2024global} & \textcolor{darkgreen}{\faTruck} \textcolor{darkgreen}{\faPhone} & 9 \\
\hline

\multirow[c]{2}{1.6cm}[-1.5em]{\centering China} & West China Hospital Mobile Lung Cancer Screening Program & Mobile LDCT units screen underserved communities. Images are transferred to West China Hospital where radiologists perform remote reporting supported by AI pulmonary nodule detection. This is the clearest published integration of mobile CT + telemedicine + AI. & Tao \textit{et al.}, 2024 \cite{tao2024telemedicine} & \textcolor{darkgreen}{\faTruck} \textcolor{darkgreen}{\faPhone} \textcolor{darkgreen}{\faMicrochip} & 9 \\
\cline{2-6}
 & Sichuan Mobile Stroke Unit & Mobile Stroke Unit equipped with CT scanner and telemedicine. Published programmes also report integration with AI-assisted image analysis in some deployments. & Kim \textit{et al.} \cite{kim2024global} & \textcolor{darkgreen}{\faTruck} \textcolor{darkgreen}{\faPhone} \textcolor{orange}{\faMicrochip} & 8-9 \\
\hline

England & East of England / Southend MSU (Ipswich) & CT-equipped ambulance delivering prehospital stroke diagnosis and treatment. Led by Prof Iris Grunwald. & \href{https://www.thetimes.com/uk/scotland/article/mobile-stroke-units-scotland-iris-grunwald-wksvqdrjg}{The Times} \cite{Times2019Stroke} & \textcolor{darkgreen}{\faTruck} \textcolor{darkgreen}{\faPhone} & 8-9 \\
\hline

\multirow[c]{2}{1.6cm}[-1em]{\centering Germany} & STEMO (Berlin) & World's first Mobile Stroke Unit (MSU). Ambulance equipped with head CT, point-of-care laboratory, neurologist, radiographer and telemedicine support. Multiple units operating in Berlin. & \href{https://stemo-info.de/}{STEMO} \cite{STEMO2011} & \textcolor{darkgreen}{\faTruck} \textcolor{darkgreen}{\faPhone} & 9 \\
\cline{2-6}
 & Hamburg MSU & Operational MSU using onboard CT and stroke specialists for prehospital diagnosis and thrombolysis. & Kim \textit{et al.} \cite{kim2024global} & \textcolor{darkgreen}{\faTruck} \textcolor{darkgreen}{\faPhone} & 9 \\
\hline

\end{tabular}
}
\end{table*}

\begin{table*}[pt!]
\ContinuedFloat
\centering
\small
\setlength{\tabcolsep}{3pt}
\renewcommand{\arraystretch}{1.1}

\caption{Continued Table 2 from previous page.}

\makebox[\textwidth][c]{%
\begin{tabular}{
    C{1.6cm}|
    C{3cm}|
    C{8cm}|
    C{2cm}|
    C{1.5cm}|
    C{1.5cm}
}
\hline
Country & Program & Description & Source & Technology & Estimated TRL \\
\hline

\multirow[c]{2}{1.6cm}[-1em]{\centering India} & Assam MSU & CT-equipped Mobile Stroke Unit supporting rural stroke diagnosis and treatment. & Kim \textit{et al.} \cite{kim2024global} & \textcolor{darkgreen}{\faTruck} \textcolor{darkgreen}{\faPhone} & 9 \\
\cline{2-6}
 & Coimbatore MSU & Operational Mobile Stroke Unit with onboard CT for acute stroke care. & Kim \textit{et al.} \cite{kim2024global} & \textcolor{darkgreen}{\faTruck} \textcolor{darkgreen}{\faPhone} & 9 \\
\hline

Nigeria & Imo MSU & First reported Mobile Stroke Unit in Africa, listed in the PRESTO registry. & Kim \textit{et al.} \cite{kim2024global} & \textcolor{darkgreen}{\faTruck} \textcolor{darkgreen}{\faPhone} & 8-9 \\
\hline

Norway & Drobak MSU & Mobile Stroke Unit serving the Oslo region with onboard CT and telemedicine-supported stroke diagnosis. & Kim \textit{et al.} \cite{kim2024global} & \textcolor{darkgreen}{\faTruck} \textcolor{darkgreen}{\faPhone} & 9 \\
\hline

Thailand & Bangkok MSU & CT-equipped Mobile Stroke Unit integrated with stroke network for rapid diagnosis and thrombolysis. & Kim \textit{et al.} \cite{kim2024global} & \textcolor{darkgreen}{\faTruck} \textcolor{darkgreen}{\faPhone} & 9 \\
\hline

Turkey & Eastern Anatolia Mobile CT Field-Hospital Service & Trailer-based 128-slice CT scanner deployed in a military field hospital. It supported brain, thoracic, abdominal–pelvic, spinal, extremity, and other CT examinations. No embedded telehealth or AI was reported. & Dag \textit{et al.}, 2022 \cite{dag2022mobile} & \textcolor{darkgreen}{\faTruck} & 8 \\
\hline

UK & NHS Mobile and Relocatable CT Services & Multiple NHS trusts use full-size CT scanners installed in mobile or relocatable trailers to expand diagnostic capacity, support Community Diagnostic Centres, manage scanner replacement, and reduce waiting lists. Remote reporting and AI integration vary between trusts. & \href{https://www.inhealthgroup.com/mobiles/}{NHS} \cite{InHealth2026Our} Illemann \textit{et al.}, 2024 \cite{illemann2024mobile} & \textcolor{darkgreen}{\faTruck} \textcolor{orange}{\faPhone} \textcolor{orange}{\faMicrochip} & 9 \\
\hline

\multirow[c]{2}{1.6cm}[-6em]{\centering US} & Levine Cancer Institute Mobile Lung Screening & Mobile LDCT service delivering screening to rural and underserved communities in North and South Carolina. Patient navigation and referral pathways were described; AI and telehealth were not. & Raghavan \textit{et al.}, 2020 \cite{raghavan2020initial} & \textcolor{darkgreen}{\faTruck} & 9 \\
\cline{2-6}
 & Wellmont Mobile Lung Screening Program & Trailer-based LDCT programme serving Appalachian rural populations. Focused on improving access to screening; no embedded AI or telehealth reported. & Headrick \textit{et al.}, 2020 \cite{headrick2020mobile} & \textcolor{darkgreen}{\faTruck} & 9 \\
\cline{2-6}
 & Houston MSU (UTHealth) & One of the earliest MSUs in North America. Equipped with onboard CT, telemedicine, laboratory and stroke neurologists. Numerous similar programmes now operate across the US. & Kim \textit{et al.} \cite{kim2024global} & \textcolor{darkgreen}{\faTruck} \textcolor{darkgreen}{\faPhone} & 9 \\
\cline{2-6}
 & Cleveland Clinic MSU & Prehospital stroke ambulance with CT scanner, laboratory and specialist team. & Kim \textit{et al.} \cite{kim2024global} & \textcolor{darkgreen}{\faTruck} \textcolor{darkgreen}{\faPhone} & 9 \\
\cline{2-6}
 & UCLA Mobile Stroke Unit & CT-equipped ambulance providing hospital-grade brain imaging, thrombolysis and telemedicine-supported stroke care. & Prabhakaran \textit{et al.} \cite{prabhakaran20262026} & \textcolor{darkgreen}{\faTruck} \textcolor{darkgreen}{\faPhone} & 9 \\
\hline

International & Siemens Mobile Lung Screening Solution & Commercial trailer containing a conventional CT scanner for mobile lung screening. Platform only; telehealth and AI depend on the local implementation. & \href{https://www.siemens-healthineers.com/en-uk/clinical-specialities/oncology/cancer-types/lung-health-check}{Siemens Healthineers} \cite{Siemens2026lung} & \textcolor{darkgreen}{\faTruck} \textcolor{orange}{\faPhone} \textcolor{orange}{\faMicrochip} & 8-9 \\
\hline

\end{tabular}
}
\end{table*}
\begin{table*}[p]
\centering
\small
\setlength{\tabcolsep}{3pt}
\renewcommand{\arraystretch}{1.1}

\caption{Regulatory-authorised and clinically deployed AI systems for CT (transferable to mobile CT) with following functions: D = Screening/Diagnosis; M = Monitoring; P = Prediction; I = Intervention/Therapeutic decision support.}
\label{tab:ai_for_ct}

\makebox[\textwidth][c]{%
\begin{tabular}{
    C{1.6cm}|
    C{2.4cm}|
    C{1.4cm}|
    C{3.9cm}|
    C{3.9cm}|
    C{3.9cm}|
    C{1.5cm}
}
\hline
Regulatory / deployment region & Product / System & Source & Function / Application & Status & Deployment & Estimated TRL \\
\hline

\multirow[c]{6}{1.6cm}[-12em]{\centering US, UK, Europe} & Brainomix e-Stroke / Brainomix 360 Stroke & \href{https://www.nice.org.uk/guidance/htg708}{NICE} \cite{Nice2024Artificial} & \textbf{D,P,I}\par NCCT, CTA and CTP for ASPECTS, LVO, perfusion and stroke treatment selection & FDA-authorised modules; CE-marked. NICE permits e-Stroke to be used in the NHS with evidence generation and professional review. & NICE reports that the recommended stroke-AI systems are already widely used in the NHS. & 9 \\
\cline{2-7}

 & RapidAI & \href{https://www.fda.gov/medical-devices/software-medical-device-samd/artificial-intelligence-enabled-medical-devices}{FDA} \cite{FDA2026AIList} & \textbf{D,P,I}\par LVO, ICH, CTP, ASPECTS and pulmonary embolism & Multiple FDA-authorised applications; CE-marked products. NICE permits its use for suspected stroke with evidence generation. & Documented use in NHS stroke pathways and other stroke centres. & 9 \\
\cline{2-7}

 & Viz.ai / Viz & \href{https://www.nice.org.uk/guidance/htg708}{NICE} \cite{Nice2024Artificial} & \textbf{D,I}\par LVO, ICH, CTA/CTP analysis and stroke-team notification & FDA De Novo/510(k)-authorised modules; appropriate CE status recognised by NICE. & NICE permits use in the NHS with evidence generation; deployed in connected stroke pathways. & 9 \\
\cline{2-7}

 & Aidoc aiOS / BriefCase & \href{https://www.fda.gov/medical-devices/software-medical-device-samd/artificial-intelligence-enabled-medical-devices}{FDA} \cite{FDA2026AIList} & \textbf{D,I}\par ICH, LVO, PE, incidental PE, fractures, pneumothorax, aortic and abdominal findings & Multiple FDA-authorised CT applications; CE-marked products. NICE places Aidoc stroke software in the research-only group. & Enterprise deployment is documented commercially, but NICE judged stroke-specific NHS evidence insufficient for routine recommendation. & 9 \\
\cline{2-7}

 & Avicenna.AI CINA Head / CINA suite & \href{https://www.fda.gov/medical-devices/software-medical-device-samd/artificial-intelligence-enabled-medical-devices}{FDA} \cite{FDA2026AIList} & \textbf{D,I}\par ICH, LVO, ASPECTS, PE, incidental PE and aortic disease & Multiple FDA 510(k)-authorised applications; CE-marked products. CINA Head is research-only under NICE stroke guidance. & Commercially deployed, but UK stroke use requires more evidence. & 8-9 \\
\cline{2-7}

 & Qure.ai qER & \href{https://www.fda.gov/medical-devices/software-medical-device-samd/artificial-intelligence-enabled-medical-devices}{FDA} \cite{FDA2026AIList} & \textbf{D,I}\par Emergency head CT, including suspected ICH, mass effect and fracture-related findings & FDA-authorised modules and CE-marked products. NICE places qER in the research-only group for stroke. & Documented clinical deployment internationally, but NICE did not recommend routine NHS stroke use outside research. & 8-9 \\
\hline

\multirow[c]{4}{1.6cm}[-4em]{\centering US, Europe} & Qure.ai qCT / qCT LN Quant & \href{https://www.fda.gov/medical-devices/software-medical-device-samd/artificial-intelligence-enabled-medical-devices}{FDA} \cite{FDA2026AIList} & \textbf{D,M}\par Pulmonary-nodule detection, measurement and longitudinal volumetry & FDA-authorised chest-CT applications; CE-marked versions. & Commercial use in chest-imaging workflows. & 9 \\
\cline{2-7}

 & Riverain ClearRead CT & \href{https://www.fda.gov/medical-devices/software-medical-device-samd/artificial-intelligence-enabled-medical-devices}{FDA} \cite{FDA2026AIList} & \textbf{D}\par Vessel suppression and pulmonary-nodule detection & FDA-authorised; CE-marked products. & Clinically deployed as an adjunctive chest-CT reading tool. & 9 \\
\cline{2-7}

 & HeartFlow FFRCT & \href{https://www.fda.gov/medical-devices/software-medical-device-samd/artificial-intelligence-enabled-medical-devices}{FDA} \cite{FDA2026AIList} & \textbf{D,P,I}\par Functional assessment of coronary stenosis from coronary CTA & FDA-authorised; CE-marked. & Established connected clinical service in coronary-CTA pathways. & 9\\
\cline{2-7}

 & MeVis Liver Suite & \href{https://www.fda.gov/medical-devices/software-medical-device-samd/artificial-intelligence-enabled-medical-devices}{FDA} \cite{FDA2026AIList} & \textbf{D,I}\par Liver segmentation, vascular analysis, volumetry and surgical planning & FDA-authorised and available in other markets. & Used in specialist liver-planning workflows. & 9 \\
\hline

South Korea, US, Europe & Coreline Soft AVIEW LCS / AVIEW Lung Nodule & \href{https://www.fda.gov/medical-devices/software-medical-device-samd/artificial-intelligence-enabled-medical-devices}{FDA} \cite{FDA2026AIList} & \textbf{D,M}\par Lung-nodule detection, measurement, emphysema and quantitative chest CT & FDA-authorised modules; CE-marked products. & Commercially used in lung-screening and chest-CT workflows. & 9 \\
\hline

\end{tabular}
}
\end{table*}

\begin{table*}[p]
\ContinuedFloat
\centering
\small
\setlength{\tabcolsep}{3pt}
\renewcommand{\arraystretch}{1.1}

\caption{Continued Table 3 from previous page.}

\makebox[\textwidth][c]{%
\begin{tabular}{
    C{1.6cm}|
    C{2.4cm}|
    C{1.4cm}|
    C{3.9cm}|
    C{3.9cm}|
    C{3.9cm}|
    C{1.5cm}
}
\hline
Regulatory / deployment region & Product / System & Source & Function / Application & Status & Deployment & Estimated TRL \\
\hline

\multirow[c]{2}{1.6cm}[-1.5em]{\centering US, Europe, Australia} & Siemens AI-Rad Companion & \href{https://www.fda.gov/medical-devices/software-medical-device-samd/artificial-intelligence-enabled-medical-devices}{FDA} \cite{FDA2026AIList} & \textbf{D,M,I}\par Chest CT quantification, organ measurements and radiotherapy contouring & Various modules have FDA authorisation, CE marking and local registrations. & Deployed within Siemens imaging ecosystems. & 9 \\
\cline{2-7}

 & RaySearch RayStation & \href{https://www.tga.gov.au/products/medical-devices/software-and-artificial-intelligence-ai/manufacturing/artificial-intelligence-ai-and-medical-device-software-regulation/ai-enabled-medical-devices-artg}{ARTG} \cite{ARTG2026AI} & \textbf{I}\par AI-assisted segmentation, radiotherapy and ablation planning using CT & FDA/CE availability; included in Australia’s ARTG. & Routine clinical use in radiotherapy centres. & 9 \\
\hline

Australia, Europe & MVision AI radiotherapy software & \href{https://www.tga.gov.au/products/medical-devices/software-and-artificial-intelligence-ai/manufacturing/artificial-intelligence-ai-and-medical-device-software-regulation/ai-enabled-medical-devices-artg}{ARTG} & \textbf{I}\par Automated organ and target contouring from planning CT & ARTG-included AI-enabled medical device; CE-marked products. & Used in radiotherapy-planning workflows. & 9 \\
\hline

\multirow[c]{3}{1.6cm}[-2em]{\centering China} & Deepwise CT image-aided detection software for intracranial aneurysms & \href{https://english.nmpa.gov.cn/2024-03/01/c_1049725.htm}{China NMPA} \cite{National2024CT} & \textbf{D}\par Detects intracranial aneurysms of at least 3 mm on head-and-neck CTA & Approved for marketing by China’s NMPA in 2024. & Commercial clinical deployment is possible following authorisation; exact national site numbers were not publicly confirmed. & 8 \\
\cline{2-7}

 & West China Hospital remote pulmonary-nodule AI system & Tao \textit{et al.} \cite{tao2024telemedicine} & \textbf{D,P}\par Lung-nodule detection and risk assessment on mobile LDCT & The publication documents clinical operation; the precise product name and NMPA registration were not reported in the paper. & Yes—documented operational use from 2020 onward in an underserved-community screening programme. & 8-9 \\
\hline

UK & NHS stroke-AI research deployments & \href{https://www.nice.org.uk/guidance/htg708}{NICE} \cite{Nice2024Artificial} & \textbf{-}\par Accipio, Aidoc, BioMind, BrainScan CT, Cercare Perfusion, CINA Head, CT Perfusion 4D, icobrain ct, Neuro Solution and qER & Assessed by NICE; research-only recommendation. & Research/limited deployment. & 7 \\
\hline

\multirow[c]{4}{1.6cm}[-3.5em]{\centering US} & Cleerly & \href{https://www.fda.gov/medical-devices/software-medical-device-samd/artificial-intelligence-enabled-medical-devices}{FDA} \cite{FDA2026AIList} & \textbf{D,P}\par Coronary plaque quantification and risk assessment from coronary CTA & FDA-authorised software. & Deployed in connected cardiac-CT workflows. & 9 \\
\cline{2-7}

 & V5med Lung AI & \href{https://www.fda.gov/medical-devices/software-medical-device-samd/artificial-intelligence-enabled-medical-devices}{FDA} \cite{FDA2026AIList} & \textbf{D}\par Pulmonary-nodule detection on chest CT & FDA 510(k)-authorised. & Commercial availability; extent of routine deployment is not fully reported publicly. & 8 \\
\cline{2-7}

 & Nanox.AI HealthFLD & \href{https://www.fda.gov/medical-devices/software-medical-device-samd/artificial-intelligence-enabled-medical-devices}{FDA} \cite{FDA2026AIList} & \textbf{D,P}\par Liver attenuation and fatty-liver-related quantitative assessment & FDA-authorised. & Commercial deployment for opportunistic CT analysis. 8-9 \\
\cline{2-7}

 & Nanox.AI HealthOST & \href{https://www.fda.gov/medical-devices/software-medical-device-samd/artificial-intelligence-enabled-medical-devices}{FDA} \cite{FDA2026AIList} & \textbf{D,P}\par Opportunistic osteoporosis assessment from existing CT & FDA-authorised. & Commercial deployment for secondary analysis of CT studies. & 8-9 \\
\hline

UK, Wales & Planned Welsh lung-cancer screening AI procurement & \href{https://www.find-tender.service.gov.uk/Notice/054418-2025}{Public Health Wales NHS Trust} \cite{Wales2025Artificial} & \textbf{D,M}\par Computer-aided detection and volumetry for LDCT lung screening & Public Health Wales commenced market engagement for a future national screening requirement; this is procurement planning rather than current routine deployment. & Planning - not included in the primary maturity analysis. & 6-7 \\
\hline

\end{tabular}
}
\end{table*}

\section{Current Technologies and Clinical Deployment}
Mobile CT, telehealth and AI represent three complementary approaches for improving access to CT-based healthcare. Mobile CT primarily addresses geographical and infrastructure barriers by bringing medical imaging services to patients in underserved areas, whereas telehealth enables images and clinical information to be accessed by specialists at distant locations. AI provides an additional layer of automated image analysis, prioritisation, quantification, and clinical decision support. Although each technology has reached relatively high maturity independently, their levels of integration and clinical deployment vary substantially. This section reviews the current deployment of these technologies, starting from mobile CT services (Section \ref{subsec:mobile_ct}), followed by telehealth-supported CT workflows (Section \ref{subsec:telehealth}) and clinically deployed and regulatory-authorised CT-AI systems (Section \ref{subsec:AI_CT}). We pay particular attention to their relevance to RRR settings, we also give particular focus on the integration of these technologies into a connected CT service (Section \ref{subsec:integration}).

\subsection{Mobile CT}
\label{subsec:mobile_ct}
Mobile CT extends conventional CT imaging beyond fixed hospital radiology rooms by installing CT equipment and relevant services within trucks, trailers, ambulances, or other transportable clinical environments. As summarised in Table~\ref{tab:mobile_CT}, operational mobile CT services have been reported across Australia, North and South America, Europe, Asia, and Africa, demonstrating that mobile CT is no longer limited to experimental deployment. However, these services differ substantially in many aspects such as their equipment and objectives.

\subsubsection{Rural and remote diagnostic and screening services}
One major deployment model uses full diagnostic or low-dose CT systems installed in transportable trucks or trailers to provide imaging directly to geographically underserved populations. In Australia, the Western NSW Remote Mobile CT Service uses a full-size CT scanner installed in a purpose-built truck to serve Bourke, Walgett, Cobar, and their surrounding rural communities. Acquired images are uploaded to the NSW imaging network for reporting, allowing mobile image acquisition to operate within the broader health-service imaging infrastructure \cite{eHealth2026Mobile}. Heart of Australia similarly operates mobile CT capability for rural and remote communities, while the Australian National Lung Cancer Screening Program is deploying mobile low-dose CT services to improve screening access for populations who face geographical barriers to fixed imaging facilities \cite{Heart2025Heart,Health2026How}. These Australian programs demonstrate the feasibility of bringing clinically established CT technology directly to RRR populations rather than requiring patients to travel to metropolitan or regional imaging centres.

Comparable models have been implemented internationally. Canada has operated publicly funded mobile CT trailers serving multiple municipalities and First Nations communities, while provincial mobile CT trailers can also maintain imaging access during planned or unplanned scanner downtime \cite{CADTH2021Canadian,PHSA2026Provincial}. In Brazil, mobile low-dose CT has been incorporated into lung cancer screening for underserved populations \cite{chiarantano2022implementation}. In the United States, mobile lung screening programs have similarly extended low-dose CT access to underserved communities \cite{raghavan2020initial,headrick2020mobile}. These programs demonstrate the important role for mobile CT: providing routine diagnostic capacity where permanent CT infrastructure is unavailable or difficult to deliver.


\subsubsection{Mobile stroke units}
A second major model is the MSU, in which a dedicated ambulance incorporates head CT, point-of-care testing, stroke expertise, and optional telemedicine service. Operational MSUs have been reported in Germany, Australia, Canada, the United States, Norway, India, Thailand, China, Brazil, Argentina, Nigeria, and the United Kingdom, among other settings (Table~\ref{tab:mobile_CT}). Unlike general mobile CT services, MSUs are designed specifically to move CT-based stroke diagnosis into mobile settings, allowing hemorrhagic and ischaemic stroke to be promptly detected and differentiated before arriving hospital and enabling earlier treatment decisions.

The Berlin STEMO program was among the earliest established MSU models, combining onboard head CT, point-of-care laboratory testing, specialist expertise, and telemedicine \cite{STEMO2011}. Similar models were then developed internationally, including the Melbourne Mobile Stroke Unit \cite{Stroke2025Stroke}, Edmonton Stroke Ambulance \cite{kim2024global}, Houston MSU \cite{kim2024global}, Cleveland Clinic MSU \cite{kim2024global}, and UCLA Mobile Stroke Unit \cite{prabhakaran20262026}. Evidence shows that MSUs are more mature and more widely deployed than many other mobile CT applications. Systematic reviews and meta-analyses have reported shorter treatment times and increased use of intravenous thrombolysis compared with conventional emergency medical services, with evidence also suggesting improved functional outcomes for patients \cite{Fatima2020MSU,Turc2022MSU}.

MSUs are particularly relevant to the broader mobile CT model because they demonstrate that mobile medical imaging services can be really deployed in RRR areas and benefit populations there. However, most MSUs remain specialised for acute neurological imaging and cannot be directly generalised to full-body mobile CT services. These MSUs therefore provide valuable operational experience for the future development of connected mobile imaging systems.

\subsubsection{Current maturity and limitations of mobile CT}
The evidence summarised in Table~\ref{tab:mobile_CT} demonstrates that mobile CT is a mature technology, with many operational services reaching an estimated TRL of 8 or above. Full-size mobile CT services demonstrate that diagnostic CT can be delivered outside fixed hospitals, while MSUs further demonstrate the feasibility of CT-based diagnosis and treatment decision-making in prehospital environments. However, mobile CT primarily addresses access to image acquisition rather than the complete diagnostic pathway. Images still require timely interpretation, reporting, communication, and integration into clinical workflow, which may remain challenging in current RRR settings with limited specialists available. These limitations provide the rationale for combining mobile CT with telehealth and AI, which are reviewed in the following sections.

\subsection{Telehealth for CT}
\label{subsec:telehealth}
Telehealth is an important technology for extending CT-based diagnostic services beyond locations where radiologists and other specialists are physically available. Within medical imaging domain, the most established model is teleradiology, in which CT images and associated clinical information are securely transmitted from the acquisition site to a radiologist at another location (e.g., urban area) for interpretation, diagnostics, and reporting. Telehealth can additionally support communication between geographically separated patients and healthcare services. The WHO defines telemedicine as synchronous or asynchronous clinical care delivered at a distance and recognises diagnostic services such as radiology within the broader scope of telehealth \cite{WHO2025Telemedicine}. In 2026, WHO Member States also endorsed strengthening access to diagnostic imaging through teleradiology, particularly for remote and underserved populations \cite{WHO2026Teleradiology}.

Teleradiology is already a mature component of contemporary radiology practice. Digital CT examinations can be transferred through Picture Archiving and Communication Systems (PACS), radiology information systems, or secure cloud-based infrastructure for remote interpretation. This model allows a CT examination acquired at a RRR facility to be interpreted by radiologists located at larger regional or metropolitan centres, including outside routine working hours. The American College of Radiology (ACR) recognises teleradiology as an established mechanism for providing timely radiological interpretation to healthcare facilities without access to an onsite radiologist \cite{Silva2013Teleradiology}. Such capability is particularly relevant to RRR healthcare, where geographical distance and shortages of specialist personnel may otherwise limit the value of locally available imaging equipment. More broadly, systematic evidence from RRR Australia indicates that telehealth can improve access to healthcare \cite{Bradford2016Telehealth}.

The role of telehealth extends beyond remote image reporting when CT findings require immediate specialist assessment and treatment decision-making. Acute stroke provides one of the clearest examples. In telestroke networks, CT and CT angiography can be reviewed remotely together with clinical information, allowing stroke specialists to support diagnosis and treatment decisions regarding transfer to comprehensive stroke centres. MSUs extend this model into the prehospital environment by combining onboard CT with telecommunication systems that connect ambulance-based teams with remote neurologists. Studies have demonstrated high agreement between remote and onsite neurologists for neurological assessment and treatment decisions, supporting the feasibility of telemedicine-assisted prehospital stroke management \cite{geisler2019telemedicine}. Therefore, mobile network performance remains important: evaluations comparing 3G and 4G connections demonstrated reliable remote neurological assessment when connectivity was available, while also identifying incomplete network coverage as a potential limitation \cite{winter20194g}.

Telehealth is therefore particularly complementary to mobile CT. Mobile CT can overcome the geographical barrier to image acquisition, whereas teleradiology and telemedicine can overcome part of the geographical barrier to specialist interpretation and clinical expertise. This relationship is evident in several mobile CT programs summarised in Table~\ref{tab:mobile_CT}, particularly MSUs in which onboard CT is combined with remote neurological assessment. Emerging approaches to remote scanner operation may further extend this concept. The ACR has recognised that remote CT and MRI scanning technologies have the potential to extend skilled technologist capability to rural and underserved locations, while emphasising the need for appropriate guideline in this emerging workflow \cite{ACR2025RemoteScanning}.

Despite its maturity, telehealth does not eliminate all barriers to CT access. Effective implementation requires reliable network connectivity, secure transmission of large imaging datasets, and appropriate cybersecurity setup. These issues may be particularly challenging in remote areas where digital connectivity and technical support are limited. Telehealth also primarily facilitates access to human expertise; it does not itself automate image interpretation or prioritisation. AI therefore represents a complementary technology that may provide rapid automated analysis before or alongside remote specialist review. The maturity and current clinical deployment of AI for CT are considered in the following section.

\subsection{Artificial Intelligence for CT}
\label{subsec:AI_CT}
AI has become an increasingly mature component of CT-based healthcare nowadays. Unlike mobile CT, which primarily addresses geographical access to image acquisition, AI has the potential to support the subsequent interpretation and decision-making by deploying either locally or through cloud-based server. This capability is particularly relevant to RRR settings, where timely access to radiologists and other specialists remains limited (even through telehealth pathway). The regulatory landscape demonstrates that CT-AI has progressed substantially beyond experimental research. The current FDA and TGA list includes, for example, AI-enabled software supporting radiotherapy planning and related image-analysis functions, although the those organisations note that its published AI list does not necessarily capture every AI-enabled device included in the regulatory list.

As summarised in Table~\ref{tab:ai_for_ct}, neurological CT is one of the most developed areas of clinical AI. Systems such as Brainomix, RapidAI, and Viz provide automated analysis of non-contrast CT (NCCT), CT angiography (CTA), and/or CT perfusion (CTP) to support detection and assessment of intracranial haemorrhage, large-vessel occlusion, early ischaemic change, perfusion abnormalities, and stroke treatment selection. Multiple modules from these systems have received FDA marketing authorisation; for example, FDA records include Brainomix 360 e-CTP, e-CTA, Triage ICH and Triage Stroke, as well as Viz ICH and Viz LVO ContaCT \cite{FDA2026AIList}. Importantly, regulatory authorisation has been accompanied by clinical implementation. The National Institute for Health and Care Excellence (NICE) currently permits e-Stroke, RapidAI, and Viz to be used within the UK National Health Service (NHS) for suspected stroke while further evidence is generated, provided that healthcare professionals continue to review the scans and existing reporting protocols are maintained \cite{NICE2024StrokeAI}.

CT-AI applications extend well beyond neurological imaging. In thoracic imaging, regulatory-authorised and clinically deployed systems support pulmonary nodule detection and measurement, lung cancer screening, pulmonary embolism detection, and quantitative assessment of emphysema and other chest abnormalities. Cardiovascular applications include coronary plaque quantification and functional assessment of coronary stenosis from coronary CTA, while abdominal applications include liver segmentation, volumetry, and quantitative tissue assessment. AI is also increasingly incorporated into interventional and therapeutic workflows, particularly automated segmentation and radiotherapy planning.

The evidence also demonstrates an important distinction between regulatory authorisation and clinical deployment. Regulatory authorisation indicates that a device has satisfied the applicable requirements for its specified intended use, but does not necessarily establish effectiveness across every patient population, scanner, institution, or clinical environment. Therefore, regulatory status, clinical evidence, and documented real-world deployment should be considered together when assessing technological maturity.

Based on these factors, many established CT-AI systems identified in this review were estimated to have reached TRL 8--9, particularly where regulatory authorisation is accompanied by routine clinical deployment (Table~\ref{tab:ai_for_ct}). This indicates that AI for conventional CT is itself relatively mature. However, this maturity does not mean that the same performance can be assumed when software is transferred to mobile CT. Differences in scanner hardware, acquisition and reconstruction protocols, image quality, and patient populations may introduce domain shift. RRR settings may face additional challenges relating to stable connection when use cloud-based deployment.

Consequently, the principal translational opportunity is increasingly not the development of CT-AI in isolation, but its safe integration with mobile CT and remote specialist services. Mature telehealth enables the outputs and images to remain under specialist clinical oversight, while CT-AI systems provide a technological foundation for automated analysis in distributed imaging environments. Evidence for bringing these components together into integrated mobile CT, telehealth, and AI pathways is considered in the following section.

\subsection{Current Integration of Mobile CT, AI and telehealth}
\label{subsec:integration}
The evidence reviewed in the previous sections indicates that mobile CT, telehealth, and AI for CT have each reached relatively high levels of technological and clinical maturity, but their integration within a single clinical pathway remains considerably less common. Combining capabilities of these technologies could create a connected imaging pathway in which CT is acquired locally, analysed rapidly by AI, and reviewed by remotely located radiologists or other specialists. Such a model is particularly relevant to RRR settings, where geographical distance and shortages of specialist expertise can affect multiple stages of the diagnostic pathway.

Current mobile CT services demonstrate different levels of this integration. Many established mobile CT programs primarily combine mobile image acquisition with conventional remote reporting, while MSUs provide a more advanced model by combining onboard CT with telemedicine-supported neurological assessment and treatment decision-making. As discussed in Sections \ref{subsec:mobile_ct} and \ref{subsec:telehealth}, operational MSUs across multiple countries demonstrate that CT images and clinical information can be transmitted from a mobile environment to remote specialists to support time-critical stroke care \cite{kim2024global}. However, explicit integration of AI into these mobile workflows remains less frequently documented.

\begin{casestudybox}
    \textbf{Setting:} Resource-constrained and geographically dispersed
    regions of Western China.
    
    \textbf{Clinical application:} Lung cancer screening using mobile
    low-dose CT.
    
    \textbf{Integrated workflow:} Mobile LDCT acquisition, remote
    specialist support through telemedicine, and AI-assisted pulmonary
    nodule analysis \cite{tao2024telemedicine}.
    
    \textbf{Scale:} 28,728 participants were registered, with 19,517
    undergoing LDCT screening.
    
    \textbf{Relevance to RRR healthcare:} The program demonstrates the
    practical feasibility of integrating mobile CT, telehealth, and AI
    within a distributed clinical pathway, providing a useful precedent
    for similar models in regional, rural, and remote healthcare systems.
\end{casestudybox}

One of the clearest examples of three-way integration is the mobile lung cancer screening program associated with West China Hospital (see case study box above, and see Supplementary Table S2 for more details). In this program, mobile low-dose CT was deployed to underserved communities, with acquired images transferred for remote specialist interpretation and AI-assisted pulmonary nodule assessment \cite{tao2024telemedicine}. The program reported 28,728 registered individuals between 2020 and 2021, of whom 19,517 underwent mobile CT screening, demonstrating that the approach was implemented beyond a small feasibility study \cite{tao2024telemedicine}. Related research from rural western China has also demonstrated the feasibility of combining mobile low-dose CT with deep-learning algorithms for pulmonary nodule detection and malignancy risk assessment in resource-constrained settings \cite{shao2022deep}. These studies provide important evidence that AI-assisted interpretation can technically and clinically operate within mobile CT workflows rather than being restricted to conventional hospital-based scanners.

Nevertheless, this level of integration is not yet representative of routine mobile CT practice. As shown in Table~\ref{tab:mobile_CT}, most operational mobile CT programs identified in this review use either mobile CT alone or mobile CT combined with some form of remote reporting or telemedicine. Only a small number explicitly document AI as part of the mobile imaging pathway. Conversely, Table~\ref{tab:ai_for_ct} demonstrates that numerous CT-AI systems have reached estimated TRLs of 8--9 through regulatory authorisation and/or established clinical deployment, but these systems have predominantly been validated and implemented using conventional hospital-based CT. The principal maturity gap therefore appears to lie not in the individual technologies, but in their integration and validation as a complete mobile clinical system.

Transferring existing CT-AI systems into mobile environments also requires careful validation. Mobile and conventional CT may differ in scanner configuration, acquisition protocols, image quality, connectivity, and patient populations. AI performance established using conventional CT therefore cannot automatically be assumed to generalise to mobile CT. Integration additionally requires interoperability among the CT scanner, AI software, telehealth platform, together with appropriate cybersecurity, data governance, and regulatory compliance. Reliable connectivity is particularly important in RRR environments because both remote specialist review and cloud-based AI may depend on transmission of large CT datasets.


Overall, the available evidence suggests a progression from mature individual technologies toward increasingly integrated systems: mobile CT, mobile CT with telehealth, and conventional CT with AI are already operational at relatively high maturity, whereas mobile CT with both telehealth and AI remains comparatively less explored and less widely deployed. Future research should therefore prioritise prospective, multicentre evaluation of integrated systems in real-world RRR environments, assessing not only diagnostic performance but also reporting time, clinical decision-making, and patient outcomes. Such evidence will be essential to determine whether the technological maturity of individual components can translate into safe, scalable, and clinically beneficial integrated mobile CT services.

\begin{table*}[t]
    \centering
    \caption{Key challenges and research priorities for integrated mobile CT, telehealth, and AI deployment in RRR settings.}
    \label{tab:challenge_priority}
    \begin{tabular}{C{3cm}|C{6.5cm}|C{6.5cm}}
        \hline
        \textbf{Domain} & \textbf{Key Challenge} & \textbf{Priority for Mobile CT Integration} \\
        \hline
        AI performance & Domain shift across scanners, protocols and populations & External validation using representative mobile CT data  \\
        \hline
        Connectivity & Limited bandwidth and unstable networks in RRR settings & Edge/local processing and resilient data transfer \\
        \hline
        Interoperability & CT, PACS, AI and telehealth platforms may use different systems & Standards-based, vendor-neutral integration \\
        \hline
        Clinical workflow & Alert fatigue, automation bias and unclear responsibilities & Human-in-the-loop workflows and defined escalation pathways \\
        \hline
        Regulation & Approval may not cover new scanners or deployment environments & Intended-use-specific regulatory assessment \\
        \hline
        Cybersecurity/privacy & Distributed transfer of imaging and clinical data & Encryption, access control and auditable governance \\
        \hline
        Evidence & Limited evidence of patient-level benefit & Prospective multicentre implementation studies \\
        \hline
        Economics/equity & High mobile infrastructure and software costs & Cost-effectiveness and equitable service-planning studies \\
        \hline
    \end{tabular}
\end{table*}

\section{Discussion}

\subsection{Current Technology Readiness}
This review demonstrates that mobile CT, telehealth, and AI for CT have each progressed substantially towards clinical maturity, although their level of integration differs. Mobile CT is already operational across multiple countries and clinical settings, ranging from full-size diagnostic and screening services to specialised MSUs. Many of these systems were therefore estimated at TRL 8--9, reflecting clinical validation and established operational deployment. Telehealth and teleradiology are similarly mature and are routinely used to connect geographically separated imaging services with radiologists and other specialists. CT-based AI has also advanced rapidly, with numerous systems receiving regulatory authorisation and entering routine or targeted clinical use.

Importantly, component maturity does not necessarily translate into system-level maturity. Conventional CT combined with AI, and mobile CT combined with remote specialist communication, are relatively well established. In contrast, fewer programs integrate mobile CT, telehealth, and AI within a single end-to-end clinical pathway. The West China Hospital mobile lung cancer screening program represents an important example, combining mobile low-dose CT, remote specialist interpretation, AI-assisted pulmonary nodule assessment, and subsequent referral and follow-up \cite{tao2024telemedicine}. However, such integrated implementations remain uncommon compared with the large number of independently deployed mobile CT, telehealth, and CT-AI systems.

The major technology-readiness gap identified by this review therefore lies less in the individual components than in their integration, validation, and deployment as a complete clinical system. This distinction is particularly important when interpreting TRL. A regulatory-authorised AI system at TRL 8--9 cannot automatically be considered equally mature when transferred from conventional hospital CT to a mobile CT environment. Similarly, operational mobile CT and telehealth systems do not necessarily establish the readiness of an integrated AI-assisted workflow. System-level readiness should therefore consider the maturity of the complete pathway.

\subsection{Opportunities for RRR Healthcare}
The complementary functions of the three technologies create particular opportunities for RRR healthcare. In acute stroke, for example, mobile CT can enable prehospital differentiation of haemorrhagic and ischaemic stroke, while telemedicine connects the mobile team with stroke specialists and AI may support rapid identification of intracranial haemorrhage, large-vessel occlusion, early ischaemic change, or perfusion abnormalities. MSU experience already demonstrates the feasibility of combining mobile CT and remote specialist decision-making, providing an important foundation for further AI integration. Similar opportunities exist for other emergency applications, including pulmonary embolism, traumatic injury, and acute cardiovascular disease.

Mobile CT also has considerable potential for screening and longitudinal care, where geographical distance can reduce participation and continuity of follow-up. Mobile low-dose CT programs for lung cancer demonstrate that imaging can be brought directly to rural and underserved populations. Integration with AI could support pulmonary nodule detection and quantitative measurement, while telehealth could facilitate specialist review and follow-up without requiring every patient to travel to a tertiary centre. The same principle could potentially be extended to other CT applications, although clinical need, radiation exposure, cost-effectiveness, and appropriate validation would need to be considered separately.

The potential benefit is therefore broader than simply replacing a radiologist with AI, which is neither the objective nor the most appropriate model for current clinical deployment. Instead, AI could function as an additional layer within a human-in-the-loop distributed imaging service, supporting clinicians by identifying potentially urgent examinations, providing quantitative measurements, and assisting prioritisation while radiologists and relevant specialists retain responsibility for clinical interpretation and decision-making. This approach may allow scarce specialist expertise to be distributed more efficiently across geographically dispersed services.

\begin{figure*}[t]
    \centering
    \includegraphics[width=0.9\linewidth]{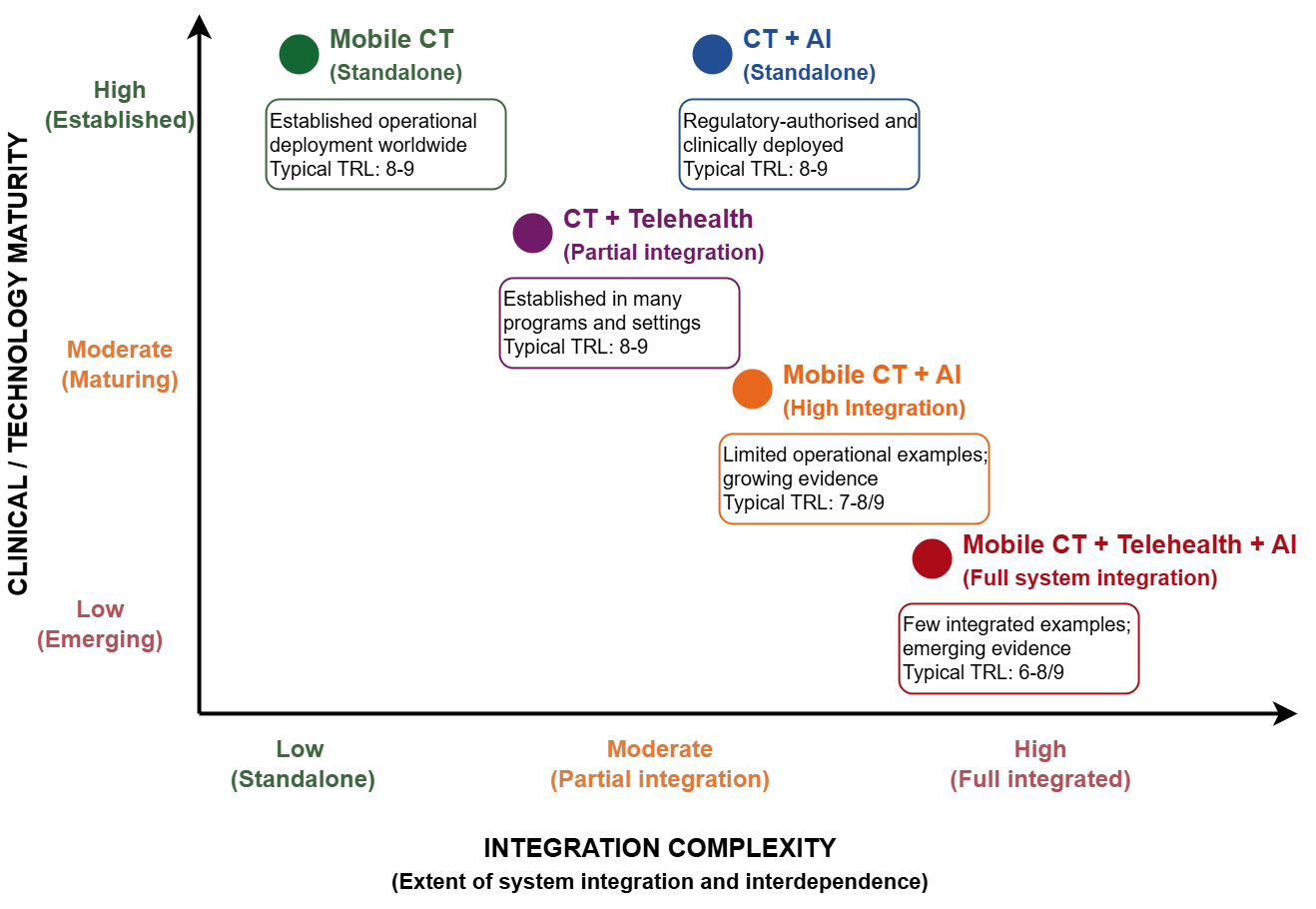}
    \caption{Relative maturity and integration of mobile CT, telehealth, and CT-based AI. The y-axis represents clinical/technology maturity, and the x-axis represents the complexity of system integration and interdependence. Technologies positioned toward the upper-left are mature as standalone solutions, whereas those toward the lower-right represent higher integration complexity with comparatively limited clinical deployment and evidence.}
    \label{fig:integration_maturity}
\end{figure*}

\subsection{Challenges and Barriers}
Despite these opportunities, several barriers must be addressed before integrated mobile CT, telehealth, and AI systems can be widely deployed. First, clinical generalisability and domain shift are important concerns. Most regulatory-authorised CT-AI systems have been developed and validated primarily using conventional hospital-based scanners, but mobile CT may differ in many aspects. Consequently, performance demonstrated on conventional CT should not be assumed to transfer directly to mobile CT environments, and external validation using representative mobile CT data is critical.

Second, integration depends on reliable digital infrastructure and interoperability. An integrated framework may involve the scanner, PACS, AI platform, cloud, and telehealth communication system. These components may be supplied by different vendors and operate under different technical standards. Reliable transfer of large CT datasets can also be challenging in locations with limited network bandwidth or unstable connectivity. Systems intended for RRR deployment should therefore consider local or edge-based processing, offline capability, data compression, redundancy, and recovery procedures in addition to cloud-based approaches.

Third, cybersecurity, privacy, governance, and regulation become increasingly complex as systems become more integrated and connected. Patient images and clinical information may move between mobile units, cloud platforms, and remote specialists. Appropriate access controls, encryption, data governance, and cybersecurity monitoring are therefore complex and essential. Regulatory authorisation may also differ across jurisdictions, and modification of an AI system or its intended deployment environment may require additional assessment. International guidance for clinical AI consequently emphasises the monitoring and transparency of the deployment \cite{Lekadir2025FutureAI,Vasey2022DECIDEAI}.

Fourth, evidence of technical performance does not necessarily demonstrate clinical and patient-level benefit. Many AI studies report sensitivity, specificity, AUC, Dice score, or reductions in image-processing time, while mobile CT studies frequently focus on feasibility and access. For integrated systems, more clinically meaningful outcomes should be systematically assessed, potential unintended consequences should also be prospectively evaluated \cite{Vasey2022DECIDEAI}.

Fifth, the strength of evidence also varies considerably across technologies. MSUs are supported by prospective trials and meta-analyses, whereas several mobile diagnostic CT programs are documented primarily through health-service or government reports. Similarly, some CT-AI systems have prospective or real-world evidence, while others are supported principally by regulatory authorisation and commercial deployment. Regulatory status, operational use, and evidence of patient-level benefit should therefore be interpreted as related but distinct indicators of clinical maturity.

Finally, the economic sustainability of mobile CT requires consideration. Mobile scanners involve vehicle or trailer infrastructure, travel costs, equipment maintenance, and ongoing software costs. Adding telehealth and AI introduces additional requirements and costs. Therefore, mobile services are unlikely to be equally appropriate for every RRR community. Their value should be assessed according to population distribution and travel distance. All above challenges and priorities are summarised into Table \ref{tab:challenge_priority}.

\subsection{Future Directions}
Future research should move beyond evaluating mobile CT, telehealth, and AI in isolation towards prospective evaluation of integrated clinical pathways. A priority is multicentre validation of regulatory-authorised or clinically mature CT-AI systems using images acquired from mobile CT scanners. Such studies should include diverse scanner models, acquisition protocols, and patient populations to determine whether performance is maintained outside the environments in which the algorithms were originally developed.

The next stage should evaluate complete mobile CT--telehealth--AI workflows in real-world RRR services. Rather than assessing AI accuracy alone, prospective studies should examine the complete pathway from image acquisition to final clinical decision-making. Evaluation should include diagnostic performance together with safety and patient outcomes, consistent with guidance for early-stage clinical evaluation of AI-supported decision-making \cite{Vasey2022DECIDEAI}.

Technical development should also prioritise interoperability and deployment resilience. Vendor-neutral interfaces based on established imaging and healthcare data standards could facilitate integration between different sides. Where connectivity is unreliable, hybrid architectures combining local AI processing with cloud-based specialist services may be preferable to systems that depend entirely on continuous high-bandwidth connections. Continuous performance monitoring will also be required to identify model drift, changes in scanner protocols, and performance differences across populations and sites.

Future systems may increasingly integrate imaging with clinical information. Rather than providing isolated image-level alerts, multimodal AI could combine CT findings with available and valuable clinical data to generate structured decision-support information for specialist review. However, increasing system complexity also increases the need for transparent outputs, appropriate human oversight, rigorous validation, and clear allocation of clinical responsibility.

Overall, the most important future direction is not necessarily the development of another independent CT-AI algorithm, but the translation of existing mature technologies into safe, interoperable, clinically validated, and sustainable RRR imaging services. Mobile CT, telehealth, and regulatory-authorised AI already provide many of the required technological components. The next challenge is to establish whether their integration can produce measurable improvements in access, timeliness, clinical decision-making, patient outcomes, and healthcare efficiency in the populations that stand to benefit most.

\subsection{Limitations of this Review}
Several limitations should be considered when interpreting this review. First, evidence on mobile CT deployment is heterogeneous and is frequently reported through government, health-service, regulatory, and manufacturer sources rather than peer-reviewed studies. Consequently, information was not uniformly available across programs. Second, regulatory and commercial landscapes for AI-enabled medical devices evolve rapidly, and the systems identified in this review represent the information publicly available at the time of the search rather than an exhaustive or permanently current inventory. Third, formal technology readiness levels were rarely reported by developers; TRLs were therefore estimated by the authors and should be interpreted as comparative indicators rather than formally assigned manufacturer TRLs. Finally, evidence for integrated mobile CT, telehealth, and AI remains limited, restricting direct comparison between integrated systems and preventing firm conclusions regarding patient-level benefit, cost-effectiveness, or generalisability across different RRR healthcare settings.

\section{Conclusion}
Mobile CT is increasingly enabling diagnostic and screening CT services to be delivered beyond conventional hospital radiology departments, including in RRR, prehospital, and other underserved settings. In parallel, telehealth has become an established mechanism for remote specialist interpretation and consultation, while numerous CT-AI systems have reached regulatory authorisation and clinical deployment. Together, these technologies support screening and diagnosis, patient monitoring, risk prediction, and intervention or therapeutic decision support.

The evidence reviewed here indicates that mobile CT, telehealth, and AI for conventional CT are individually relatively mature, whereas their integration into a complete mobile CT--telehealth--AI pathway remains less widely demonstrated. This represents both the principal technology-readiness gap and an important opportunity for RRR healthcare. Future work should prioritise prospective, multicentre validation of integrated systems in representative mobile and RRR environments, with evaluation extending beyond technical accuracy to workflow, safety, clinical outcomes, equity, cost-effectiveness, and patient and clinician acceptance. With appropriate validation, governance, interoperability, and human oversight, integration of mobile CT, telehealth, and AI has the potential to extend timely CT-based healthcare and specialist support to populations for whom geographical distance and workforce limitations continue to restrict access.

\bibliography{references}





\end{document}



\maketitle


\begin{table*}[h]
\centering
\small
\caption{Database and supplementary search strategies used to identify mobile CT, telehealth, and CT-based artificial intelligence technologies. Search syntax was adapted to the requirements of each database.}
\label{tab:supp_search_strategy}

\begin{tabular}{p{2.4cm}|p{11.8cm}|p{2.2cm}}
\hline
\textbf{Source} & \textbf{Search strategy / key search terms} & \textbf{Search date} \\
\hline

PubMed &
\texttt{("computed tomography"[Title/Abstract] OR CT[Title/Abstract]) AND ("mobile CT"[Title/Abstract] OR "mobile computed tomography"[Title/Abstract] OR "portable CT"[Title/Abstract] OR "mobile stroke unit"[Title/Abstract] OR "mobile lung screening"[Title/Abstract])}
&
31$^\text{st}$, August, 2026
\\
\hline

PubMed &
\texttt{("computed tomography"[Title/Abstract] OR CT[Title/Abstract]) AND ("telehealth"[Title/Abstract] OR "telemedicine"[Title/Abstract] OR "teleradiology"[Title/Abstract] OR "telestroke"[Title/Abstract] OR "remote interpretation"[Title/Abstract])}
&
31$^\text{st}$, August, 2026
\\
\hline

PubMed &
\texttt{("computed tomography"[Title/Abstract] OR CT[Title/Abstract]) AND ("artificial intelligence"[Title/Abstract] OR "deep learning"[Title/Abstract] OR "machine learning"[Title/Abstract] OR "computer-aided diagnosis"[Title/Abstract]) AND ("clinical deployment"[Title/Abstract] OR "clinical implementation"[Title/Abstract] OR FDA[Title/Abstract] OR "510(k)"[Title/Abstract] OR "De Novo"[Title/Abstract])}
&
31$^\text{st}$, August, 2026
\\
\hline

Scopus &
\texttt{TITLE-ABS-KEY(("computed tomography" OR CT) AND ("mobile CT" OR "mobile computed tomography" OR "portable CT" OR "mobile stroke unit" OR "mobile lung screening"))}
&
31$^\text{st}$, August, 2026
\\
\hline

Scopus &
\texttt{TITLE-ABS-KEY(("computed tomography" OR CT) AND (telehealth OR telemedicine OR teleradiology OR telestroke OR "remote interpretation"))}
&
31$^\text{st}$, August, 2026
\\
\hline

Scopus &
\texttt{TITLE-ABS-KEY(("computed tomography" OR CT) AND ("artificial intelligence" OR "deep learning" OR "machine learning" OR "computer-aided diagnosis") AND ("clinical deployment" OR "clinical implementation" OR FDA OR "510(k)" OR "De Novo" OR TGA OR ARTG))}
&
31$^\text{st}$, August, 2026
\\
\hline

Web of Science &
\texttt{TS=(("computed tomography" OR CT) AND ("mobile CT" OR "mobile computed tomography" OR "portable CT" OR "mobile stroke unit" OR "mobile lung screening"))}
&
31$^\text{st}$, August, 2026
\\
\hline

Web of Science &
\texttt{TS=(("computed tomography" OR CT) AND (telehealth OR telemedicine OR teleradiology OR telestroke OR "remote interpretation"))}
&
31$^\text{st}$, August, 2026
\\
\hline

Web of Science &
\texttt{TS=(("computed tomography" OR CT) AND ("artificial intelligence" OR "deep learning" OR "machine learning" OR "computer-aided diagnosis") AND ("clinical deployment" OR "clinical implementation" OR FDA OR "510(k)" OR TGA OR ARTG))}
&
31$^\text{st}$, August, 2026
\\
\hline

IEEE Xplore &
\texttt{("computed tomography" OR CT) AND ("mobile CT" OR "portable CT" OR "mobile stroke unit" OR telehealth OR telemedicine OR teleradiology OR "artificial intelligence" OR "deep learning")}
&
31$^\text{st}$, August, 2026
\\
\hline

Google Scholar &
\texttt{"mobile CT" OR "mobile computed tomography" OR "mobile stroke unit" AND clinical}
&
31$^\text{st}$, August, 2026
\\
\hline

Google Scholar &
\texttt{"CT" AND (telehealth OR telemedicine OR teleradiology OR telestroke) AND clinical}
&
31$^\text{st}$, August, 2026
\\
\hline

Google Scholar &
\texttt{"CT" AND "artificial intelligence" AND (FDA OR TGA OR NICE OR NMPA OR "clinical deployment")}
&
31$^\text{st}$, August, 2026
\\
\hline

FDA &
AI-enabled medical devices and 510(k)/De Novo databases searched for CT-related radiology products, including neurological, thoracic, cardiovascular, abdominal, oncological, musculoskeletal, and interventional applications.
&
31$^\text{st}$, August, 2026
\\
\hline

TGA / ARTG &
AI-enabled medical-device listings and ARTG records searched for CT-related software and image-analysis systems.
&
31$^\text{st}$, August, 2026
\\
\hline

NICE &
Health-technology guidance searched for CT-based AI, stroke imaging AI, teleradiology, and related clinically deployed decision-support systems.
&
31$^\text{st}$, August, 2026
\\
\hline

China NMPA &
Medical-device approval records searched for CT-related AI and computer-aided detection systems.
&
31$^\text{st}$, August, 2026
\\
\hline

Government / health-service websites &
Targeted searches for mobile CT, mobile stroke units, mobile lung cancer screening, rural CT, remote CT, teleradiology, and telehealth-supported CT services. Sources included national and regional health departments, health services, hospitals, and public-health agencies.
&
31$^\text{st}$, August, 2026
\\
\hline

Manufacturer websites &
Targeted searches used to confirm product availability, intended use, regulatory status, interoperability, and documented deployment where stronger independent sources were unavailable.
&
31$^\text{st}$, August, 2026
\\
\hline

Citation searching &
Reference lists and forward citations of key mobile CT, mobile stroke unit, telehealth, and CT-AI publications were screened for additional eligible technologies and clinical deployments.
&
31$^\text{st}$, August, 2026
\\
\hline

\end{tabular}
\end{table*}

\clearpage

\begin{table*}[t]
\centering
\caption{Case study of integrated mobile CT, telehealth, and AI for
lung cancer screening in Western China.}
\label{tab:supp_case_study}
\begin{tabular}{p{0.22\textwidth} p{0.70\textwidth}}
\hline
\textbf{Characteristic} & \textbf{Description} \\
\hline

Clinical setting &
Lung cancer screening in resource-constrained and geographically
dispersed regions of Western China. \\
\hline

Imaging technology &
Mobile low-dose computed tomography (LDCT) used to extend CT
screening beyond conventional hospital-based imaging facilities. \\
\hline

Telehealth component &
Imaging and clinical information were supported by remote specialist
services, enabling expertise to be accessed from geographically
distributed screening locations. \\
\hline

AI component &
AI-assisted CT analysis was incorporated to support pulmonary nodule
detection and assessment. \\
\hline

Scale &
28,728 participants were registered, of whom 19,517 underwent LDCT
screening. \\
\hline

Integrated workflow &
The program combined mobile CT acquisition, remote clinical support,
and AI-assisted image analysis within a distributed lung cancer
screening pathway. \\
\hline

Evidence of maturity &
The large-scale clinical implementation provides evidence that the
three technological components can be integrated beyond a
proof-of-concept setting. \\
\hline

Relevance to RRR healthcare &
The model provides an important precedent for extending advanced
CT-based services to populations with limited local access to imaging
infrastructure and specialist expertise. \\
\hline

Transferability considerations &
Implementation in other RRR health systems may be influenced by
differences in connectivity, workforce availability, image-transfer
infrastructure, governance, reimbursement, regulation, and existing
models of care. \\
\hline

Key implication &
The case provides relatively strong evidence of \textit{integration
feasibility}, but does not by itself establish \textit{system-level
generalisability} across different health systems and geographical
settings. \\

\hline
\end{tabular}
\end{table*}
